\documentclass[11pt]{elsarticle}
\makeatletter
\def\ps@pprintTitle{%
 \let\@oddhead\@empty
 \let\@evenhead\@empty
 \def\@oddfoot{}%
 \let\@evenfoot\@oddfoot}
\makeatother
\usepackage[utf8]{inputenc}
\usepackage{filecontents}
\usepackage{setspace}
\usepackage{natbib}
\usepackage{gensymb}
\usepackage{amssymb}
\usepackage{mathtools}
\usepackage{hyperref}
\usepackage{natbib}
\usepackage{graphicx}
\usepackage{enumitem}
\usepackage{adjustbox}
\usepackage{array}
\usepackage[section]{placeins}
\usepackage{longtable}
\usepackage{appendix}
\usepackage{import}
\usepackage{tabularx}
\usepackage{rotating}
\usepackage{tabularx}
\usepackage{ragged2e}
\usepackage{multirow}
\newcolumntype{d}[1]{D{.}{.}{#1}}        
\setcitestyle{square}
\begin{document}
\begin{frontmatter}
\title{Application of Deep Learning in Fundus Image Processing for Ophthalmic Diagnosis - A Review}
\author[1,2]{Sourya Sengupta}
\author[1,2]{Amitojdeep Singh}
\author[1,2]{Henry A. Leopold}
\author[3]{Tanmay Gulati}
\author[1,2]{Vasudevan Lakshminarayanan}
\address[1]{Theoretical and Experimental Epistemology Lab, School of Optometry  and  Vision Science,University of Waterloo, Ontario,Canada}
\address[2]{Department of System Design Engineering,University of Waterloo, Ontario,Canada}
\address[3]{Department of Computer Science and Engineering, Manipal Institute of Technology, India}

\begin{abstract}
An overview of the applications of deep learning in ophthalmic diagnosis using retinal fundus images is presented. We also review various retinal image datasets that can be used for deep learning purposes. Applications of deep learning for segmentation of optic disk,  blood vessels and retinal layer as well as detection of lesions are reviewed. Recent deep learning models for classification of diseases such as age-related macular degeneration, glaucoma, diabetic macular edema and diabetic retinopathy are also reported.
\end{abstract}
\end{frontmatter}

\section{Introduction} \label{sec:intro}
In the United States, more than 40 million people suffer from acute eye related diseases that may lead to complete vision loss if left untreated \cite{whitcher2001corneal}. Many of these diseases involve the retina. Glaucoma, diabetic retinopathy and age-related macular degeneration are some of the most common retinal diseases. Figure 1 is a fundus photograph of the retina with various structures and disease manifestations.

\begin{figure}[hbt!]
\centering
  \centering
  \includegraphics[width=.33\linewidth]{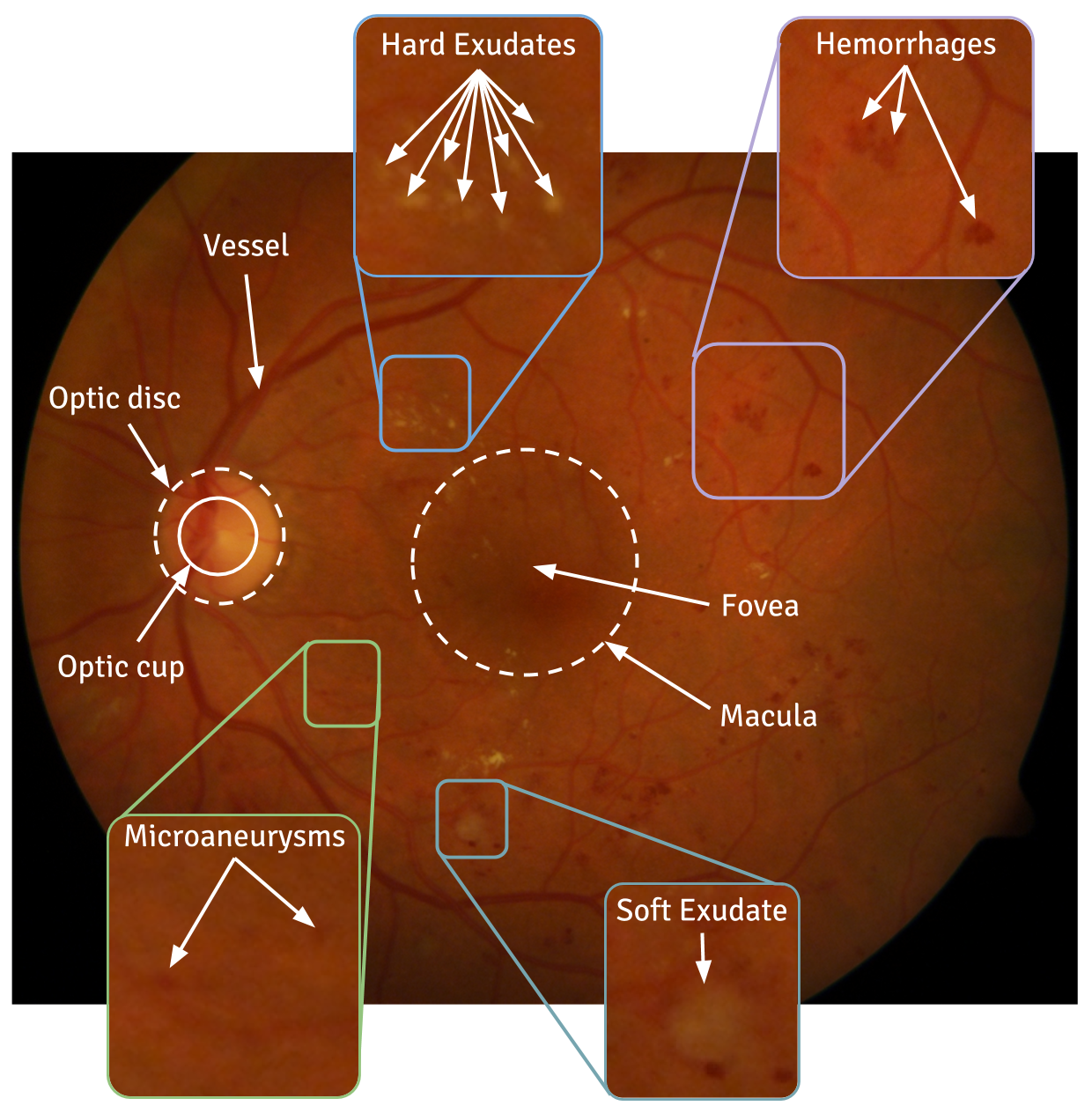}
  \caption{Fundus Photograph showing retinal morphologies and pathologies \cite{leopold_2017}.}
  \label{fig:fundus}
\end{figure}

Glaucoma is one of the major causes of blindness; it is estimated that by 2020 glaucoma will affect almost 80 million people in the world \cite{costagliola2009pharmacotherapy}. The two main types of this disease are \textit{ open-angle} glaucoma and \textit{angle closure} glaucoma. About 90\% of the affected people suffer from primary open-angle glaucoma \cite{nicolela2016optic}. Traditionally glaucoma is diagnosed by calculating what is called the optic cup to disk ratio . Neuroretinal rim loss, visual fields and retinal nerve fibre layer defects are also some of the measures used by ophthalmologists for diagnosis.
Diabetic retinopathy (DR) is another common cause of human vision loss. It is expected that the percentage of diabetic patients worldwide will increase from 2.8\% in 2000 to 4.4\% in 2030. Diabetes is quite common in persons above the age of 30; uncontrolled diabetes can lead to DR \cite{krolewski1986risk}. Early stages of DR are less severe and clincially managed.. It is characterized by various abnormalities in retina such as microaneurysms (MA) and other small lesions caused by rupture of thin retinal capillaries; these are early indicators for DR. Some of the other manifestations include hard exudates, soft exudates or cotton wool spots (CWS), hemorrhages (HEM), neovascularization (NV) and macular edema (ME) (see Figure \ref{fig:fundus}) \cite{leopold_2017}.

Age-related macular degeneration (AMD) is another common vision related problem. It can result in loss of vision in the middle of the visual field in the human eye, and with time there is a complete loss of central vision \cite{jager2008age}. In the United States, about 0.4\% people from age range 50 to 60 suffer from this disease and around 12\% people who are over 80 years old are affected \cite{friedman2004prevalence}.
Health-care in most countries suffers from a low doctor to patient ratio. Due to an overburdened patient-care system, diagnosis and proper treatment becomes error-prone and time-intensive. On the other hand, sufficient amount of data are generated everyday in various health clinics and hospitals, but it is rarely utilized for computer aided diagnostics (CAD) applications and not available publicly.
\cite{leopold_2017}.
 During the past few years, artificial intelligence algorithms have been used in classifying different types of data including images.In retinal image analysis, the traditional CAD system architectures takes several predefined templates and kernels to compare with manually annotated and segmented parts of these images. Deep learning models are extremely powerful architectures to find patterns between different nonlinear combinations of different types of data. It derives relevant necessary representations from the data without the requirement of manual feature extraction. In recent years, deep learning algorithms are replacing most of the traditional machine learning algorithms and in most of cases outperforming the traditional classifiers. General details of the different deep learning architectures like Alexnet \cite{krizhevsky2012imagenet}, VGG \cite{He_2016_CVPR}, Sparse Autoencoder \cite{ng2011sparse} can be found in  \cite{guo2016deep}.
This review focuses on the application of different deep learning architectures and algorithms for retinal fundus image processing especially for segmentation and classification problems. Table 1 gives an overview of existing fundus image datasets which are commonly used in deep learning models.
Section \ref{sec:applications} reviews various applications of deep learning for detection and diagnosis of ophthalmic diseases from retinal fundus images. Section 3 discusses several future research directions and critical insights.
\section{Application in Retinal Image Processing Techniques} \label{sec:applications}
\begin{tiny}

\begin{longtable}{m{1.5cm}m{3cm}m{3cm}m{3cm}m{2.5cm}}

\caption {Fundus Image Dataset Information}\\ 
\hline
\hline
\textbf{Dataset Name} & \textbf{Images} & \textbf{Usage} & \textbf{Camera} & \textbf{Availability} \\\\
\hline
ACHIKO-K \cite{zhang2013achiko} &  258 manually annotated images, 114 Glaucoma, 144 Normal & Glaucoma detection &  & \href{https://oar.a-star. edu.sg/jspui/handle/123456789/1080?mode=full}{Available Online} \\

\hline 
AREDS \cite{clemons2003age} & Approx. 206,500 images & AMD detection & & \href{https://www. ncbi.nlm.nih.gov/projects/gap/cgi-bin/study.cgi?study_id=phs000001.v3.p1}{Upon Request} \\\\

\hline
CHASE \cite{owen2009measuring} & 28 images & Blood vessel segmentation  & & \href{https://blogs.kingston.ac.uk/retinal/chasedb1/}{Available Online}\\\\

\hline
CLEOPATRA \cite{sivaprasad2014multicentre} & 298 images & OD segmentation & & Not Available Publicly \\\\

\hline
DIARETDB1 \cite{kauppi2007diaretdb1} & 88 images,84 DR and 4 normal & DR detection & Fundus Camera FOV 50\degree & \href{http://www.it.lut.fi/project/imageret/diaretdb1/}{Available Online} \\\\

\hline
DIARETDB0 \cite{kauppi2006diaretdb0} & 130 images, 20 normal and 110 DR & DR detection &  & \href{http://www.it.lut.fi/project/imageret/diaretdb0/}{Available Online} \\\\

\hline
DRIONS-DB \cite{carmona2008identification} & 110 images, 23.1\% Chronic Glaucoma and 76.9\% Eye Hypertensio & Glaucoma detection & & \href{http://www.ia.uned.es/~ejcarmona/DRIONS-DB.html}{Available Online} \\\\

\hline
DRISHTI-GS \cite{sivaswamy2014drishti} & 101 images & Glaucoma detection & Fundus Camera with FOV 30\degree & \href{http://cvit.iiit.ac.in/projects/mip/drishti-gs/mip-dataset2/Home.php}{Available Online} \\\\

\hline
DRIVE \cite{staal2004ridge} & 40 images,33 normal and 7 mild DR & Vessel segmentation & Canon CR5 non-mydriatic 3CCD camera with FOV 45\degree & \href{https://www.isi.uu.nl/Research/Databases/DRIVE/}{Avaliable Online} \\\\

\hline
Kaggle/ EyePACS \cite{kagg} & 35126 images& DR detection &  & \href{https://www.kaggle.com/c/diabetic-retinopathy-detection/data}{Available on Registration} \\\\

\hline
HRF \cite{budai2013robust} & 45 images,15 images each of healthy, DR, glaucomatous patients & Glaucoma detection & Canon CR-1 fundus camera with FOV 45\degree & \href{https://www5.cs.fau.de/research/data/fundus-images/}{Available Online} \\\\

\hline
KORA \cite{brandl2016features} & images from 2,840 patients & AMD detection  & & \href{https://epi.helmholtz-muenchen.de}{Available Online} \\\\

\hline
SEED \cite{zheng2013much} & 235 images,43 glaucoma and 192 normal & Glaucoma & & Not available online \\\\

\hline
STARE \cite{hoover2000locating} & 400 images,blood vessel annotation on 40 images & Vessel segmentation & TRV50 fundus camera with FOV 35\degree & \href{http://cecas.clemson.edu/~ahoover/stare/}{Available Online} \\\\

\hline
MESSIDOR \cite{decenciere2014feedback} & 1200 images & OD segmentation,Lesion detection & Color Video 3CCD camera with FOV 45\degree & \href{http://www.adcis.net/en/third-party/messidor/}{Available Online} \href{https://www.kaggle.com/google-brain/messidor2-dr-grades}{MESSIDOR-2 Upon Request} \\\\

\hline
e-optha \cite{decenciere2013teleophta} & 47 images with exudates, 35 without. 233 normal images and 148 MA images & Lesion detection & & \href{http://www.adcis.net/en/third-party/e-ophtha/}{Available Online} \\\\

\hline
ONHSD \cite{lowell2004optic} & 99 images & OD,ON segmentation & Canon CR6 45MMNf with FOV 45\degree & \href{http://www.aldiri.info/Image\%20Datasets/ONHSD.aspx}{Available Online} \\\\

\hline
ORIGA \cite{zhang2010origa} & 650 retinal images & Glaucoma detection & & Not Available Online \\\\

\hline
RIGA \cite{almazroa2018retinal} & 760 retinal fundus images & Glaucoma detection & & \href{https://deepblue.lib.umich.edu/data/concern/data_sets/3b591905z}{Available Online} \\\\

\hline
RIM-ONE \cite{fumero2011rim} & 783 images & OD segmentation & Nidek AFC-210 Can EOS 5D Mark II & Not Available Publicly \\\\

\hline
REFUGE \cite{refuge} & 1200 annotated images & Glaucoma detection & &  \href{https://refuge.grand-challenge.org}{Available Online} \\\\

\hline
Retinopathy Online Challenge \cite{niemeijer2010retinopathy} & 100 fundus images & Lesion detection & Topcon NW 100, Topcon NW 200, Canon CR5-45NM & \href{http://webeye.ophth.uiowa.edu/ROC/}{Available on Registration} \\\\
\hline

\end{longtable}
\end{tiny}
To the best of our knowledge, the very first application of computer-aided methods to  clinical ophthalmology was by Goldbaum et al. in 1994 \cite{goldbaum1994interpretation}. The authors concluded that a neural network could be trained and modelled as efficiently as a trained reader for glaucoma visual field interpretation. Another early application was the use of a neural network to predict astigmatism after cataract surgery \cite{vengupap}.

Segmentation is an important step for automatic cropping of the region of interest for further processing. An image may possess some unwanted distortions which hamper proper processing. Noise can be present in the images and the illumination may not be uniform across the image. Hence for proper visualization, different parts of an image should be segmented. Over the last few years different deep learning approaches combining with various methodologies were reported to solve segmentation problems.

In this review, we will discuss recent articles where different deep learning architectures have been implemented for ophthalmic applications with fundus images. Figure 2 shows yearwise trends of published literature and also number of papers for different application areas. It can be seen that number of published papers on deep learning for fundus imaging for ophthalmic diagnosis has increased significantly starting from 2014. In this review, published papers upto December 2018 have been reviewed. Papers were collected through search queries on google scholar with various keywords like deep learning, ophthalmology, image segmentation, classification, fundus photos, image datasets (e.g.- MESSIDOR, DRIVE, STARE, EYEPACS, RIGA etc), retina. Different performance measures like accuracy (Acc) , sensitivity (SN), specificity (SP), area under curve (AUC), F1 score, DICE Score are mentioned for different application areas.
Please refer to \cite{leopold2017pixelbnn} for details on the performance indicators discussed herein.

\begin{figure}[htb!]
\includegraphics[width=0.5\linewidth]{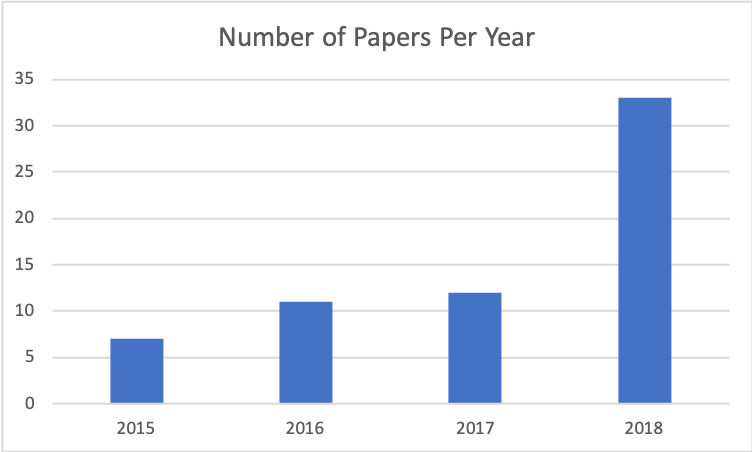}
\includegraphics[width=0.5\linewidth]{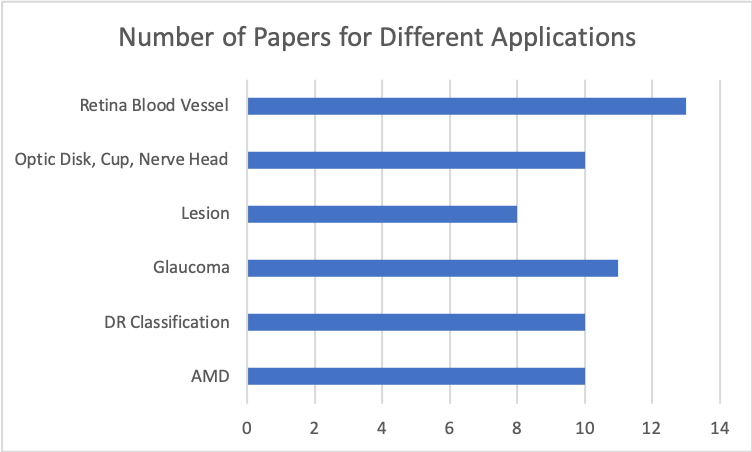}

\caption{Analysis of Reviewed Articles}
\label{fig:flowchart}
\end{figure}

\subsection{Fundus Image Applications} \label{sec:fundus_apps}

\subsubsection{Optic Disc(OD), Cup(OC) and Nerve Head Segmentation(ONH)}  \label{sec:seg_od}

The first implementation of deep learning architecture in OD segmentation was proposed by Lim et al. \cite{lim2015integrated} in 2015. During this time (2015) CNN had already been successfully implemented in various biomedical segmentation problems \cite{ciresan2012deep} \cite{cirecsan2013mitosis}. The authors developed CNN for calculating cup-to-disc ratio as a measure of the presence of glaucoma to overcome the need for hand-crafted feature extraction methods of shallow machine learning algorithms.
Since this publication, there have been significant advances in deep learning architectures. Maninis et al. \cite{maninis2016deep} experimented on fundus images to segment both blood vessels and OD together using the VGG model \cite{He_2016_CVPR} with a smaller modification of layers. Feng et al. \cite{feng2017deep} performed both OD and exudates segmentation. A fully convolutional neural network (FCN) of U-Net architecture, very popular for biomedical image segmentation problems \cite{ronneberger2015u}, was modified by replacing convolutional layers with residual blocks, (inspired by He et al. \cite{he2016deep}) and used to build a unified architecture. Sevastopolsky \cite{sevastopolsky2017optic} also used U-Net architecture(with reduced number of filters in each convolutional layer) to segment both OD and OC decreasing both time and space complexity.
In \cite{sevastopolsky2017optic} the authors segmented both OD and OC separately. Edupuganti et al. \cite{edupuganti2018automatic} implemented one shot segmentation pipeline for segmenting OD and OC for glaucoma analysis.ImageNet(\url{http://www.image-net.org/}) was used for initialization of the FCN encoder. Utilizing the concept of unified segmentation architecture like \cite{maninis2016deep} or \cite{feng2017deep} Al-Bander et al. \cite{al2018multiscale} proposed deep learning based segmentation architecture for both OD and fovea together. 
Recently Mitra et al. \cite{mitra2018region} reported some drawbacks of Al-Bander et al. \cite{al2018multiscale}, as Al-Bander et al. used grayscale images which resulted in some data loss. Their proposed architecture in \cite{al2018multiscale} utilized Dropout layers at different stages which arbitrarily dropped neurons resulting in further data loss. To address and overcome these shortcomings, \cite{mitra2018region} used batch-normalization in CNN for OD detection. More recently Liu et al. \cite{liu2018deep} used fundus images to implement deep learning based segmentation architectures to segment glaucomatous OD. A previously trained model with ImageNet database was used and the output layer was replaced by a new output layer with 2 nodes for 2 different classes- normal and glaucoma. In contrast with the previous studies, this work gathered a larger amount of data from different sources with different image qualities and resolutions. Hence this model can be considered as more robust than most of the other works. 
Sun et al. \cite{sun2018optic} employed a faster R-CNN architecture as a deep object detection architecture to segment OD from fundus. Ghassabi et al. \cite{ghassabi2018unified} introduced a consolidated approach of ONH and cup segmentation for glaucoma assessment which is effective even when there are non-obvious neuroretinal rim, peripapillary atrophy and low intensities of the optic cup as it gave a better performance (as shown by the overlapping error).
\begin{table}[hbt!]
\caption {Summary of Some Optic Disc Segmentation Results} \label{tab:title}
\begin{adjustbox}{width=1\textwidth}
\begin{tabular}{m{2.5cm}m{2.5cm}m{2.5cm}m{2.5cm}m{2cm}m{2cm}m{2cm}m{2cm}m{2cm}m{2cm}}
\hline
\textbf{Reference} & \textbf{Architecture} & \textbf{Dataset} & \textbf{Acc} & \textbf{SN} & \textbf{SP} & \textbf{AUC} & \textbf{F1 Score} &\textbf{DICE Score} & \textbf{Overlapping Error}\\
\hline
Lim et al. & 3-class CNN & MESSIDOR, SEED-DB &  &  &  & .847 \\ \hline 
Maninis et al. & CNN DRIU & DRIVE, STARE, DRIONS-DB, RIM-ONE &  &  &  &  & .822, .831, .971, .959 \\ \hline
Feng et al. & FCN & DRIONS-DB & & 93.12\% & 99.56\% & & 0.9093\\ \hline
Sevastopolsky & U-Net with lesser filters & DRIONS-DB, RIM-ONE & & & & & & .94, .95 \\ \hline
Edupuganti et al. & VGG16 FCN & Drishti-GS &  &  &  &  & .967  \\ \hline
Al-Bander et al. & & MESSIDOR & 96.89\% \\ \hline
Mitra et al. & CNN & MESSIDOR, EyePACS & 99.05\%, 98.78\% &  & 99.14\%, 98.17\% &  & \\ \hline
Sun et al. & Faster R-CNN & ORIGA Dataset & 93.1\% \\ \hline
Liu et al. & ResNet50 FCN & 3 centres from Sydney, HRF, RIM-ONE & 91.6\% & 86.7\% & 96.5\% & .97\\ \hline
Ghassabi et al. & WTA Neural Network, SOM Neural Network & Stein Eye Institute, Labbafi Nedjad hospital of Iran, RIMONE, DIARETDBO & & & & & & & 9.6\%(ONH Seg.) \& 25.1\%(Cup Seg.) \\ \hline
Tan et al. & 10-layer CNN & CLEOPATRA-DB & & 87.58\%-exudates 71.58\%-dark lesions & & \\\hline

\hline
\end{tabular}
\end{adjustbox}
\end{table}

\subsubsection{Lesion Segmentation and Detection} \label{sec:ma}
Lesion detection is an important step for DR screening. Different deep learning based studies on lesion detection and segmentation are discussed below.

Haloi \cite{haloi2015improved} was the first to implement deep neural network to detect MA for DR screening. He used a 5 layer pixel based deep neural network to detect MA. Shan et al. \cite{shan2016deep} found biological cell nuclei detection and MA detection problems quite similar and employed stacked sparse autoencoder (SSAE) proposed in a nuclei detection problem \cite{xu2016stacked}. Classification was done for MA and non-MA patches. Image patches were passed through the SSAE to obtain features and a softmax layer was used to classify the labels. The previous studies mainly addressed detection of MA, but in DR screening bigger hemorrhages are also important. Grinven et al.\cite{van2016fast} proposed a methodology to detect hemorrhages in retinal fundus images by classifying the lesions. CNN was implemented and a selective sampling algorithm was introduced to dynamically select misclassified training samples. It was found to decrease the time of epochs and also to enhance the AUC as compared for the CNN with no selective sampling.
All previous studies \cite{haloi2015improved} \cite{shan2016deep} \cite{van2016fast} tried to detect different lesions separately, Tan et al. \cite{tan2017automated} applied a 10 layer CNN to automatically segment exudates, MA and hemorrhages using a single framework. Orlando et al. \cite{orlando2018ensemble} also worked on detection of both MA and HE together using 3 different data-sets combining hand-crafted feature and deep features learned from CNN. Previously there were very few studies analyzing the effectiveness of combining two feature methods \cite{annunziata2016accelerating}. Deep learned feature vectors using 4 convolutional layers and 1 fully connected layer from CNN(trained using LeNet architecture) were created by combining handcrafted features drawn from the green channel of the normalized and equalized image. 
Lam et al. \cite{lam2018retinal} used a deep learning architecture to detect the presence of five classes of red lesions i.e. normal, microaneurysms, hemorrhages, exudates, retinal neovascularization using EyePACS. The CNN was trained with 1050 images using GoogleNet architecture \cite{guo2016deep}. Patches were extracted with varying shapes and sizes according to the size of the lesions. For testing, a sliding window was introduced to make a full scan over the whole image by the CNN to give a multiclass outcome probability.
Son et al. \cite{son2018classification} proposed a cost-effective method to localize lesions which improved precision during training by using regional annotation of findings. Badar et al. \cite{badar2018simultaneous} used an encoder-decoder based FCN architecture calculating pixel-wise segmentation of multi-class retinal pathologies (exudates, hemorrhages  \& cotton wool spots) and achieved state of the art results.
Khojasteh et al. \cite{khojasteh2018introducing} introduced an innovative framework and architecture for CNN by inserting a pre-processing layer for recognition of HE and MAs. Chudzik et al \cite{chudzik2018exudates} presented a segmentation method which utilized similar a combination of CNN and codebook structure.
\begin{table}[hbt!]
\caption {Summary of Red Lesion Detection Studies} \label{tab:title}
\begin{adjustbox}{width=1\textwidth}
\begin{tabular}{m{2.5cm}m{2.5cm}m{2.5cm}m{2.5cm}m{2.5cm}m{2.5cm}m{2.5cm}m{2.5cm}m{2.5cm}}
\hline
\textbf{Reference} & \textbf{Architecture} & \textbf{Dataset} & \textbf{Acc} & \textbf{SN} & \textbf{SP} & \textbf{AUC} \\ \hline

Haloi & 5 layer CNN & MESSIDOR, ROC & 96\% & 97\% & 96\%& .982, .98 \\ \hline
Shan et al. & Transfer Learning (SSAE) & DIARETDB &  91.38\% & & & .916\\ \hline
Grinven et al.& CNN using OxfordNet & MESSIDOR, EyePACS & & 91.9\%, 83.7\% & 91.8\%, 85.1\% & .972, .895 & \\ \hline
Tan et al. & 10 layer CNN & CLEOPATRA & 87.58\%, 71.58\% & & & \\ \hline
Orlando et al. & CNN using LeNet architecture & e-optha, MESSIDOR & & & & .8812, .8932 \\ \hline
Lam et al. & CNN using GoogleNet3 & EyePACS, e-optha & 98\% & & & .95 & \\ \hline
Son et al. & CNN with residual, reduction, avg. pooling, atrous pyramid pooling layers  & Seoul National University Bundang Hospital & & & & 0.9895\\ \hline
Badar et al. & Encoder-Decoder based FCN & MESSIDOR & 97.86\% & 80.93\% & 98.54\% \\ \hline
Khojasteh et al. & CNN with preprocessing after 1st conv layer & DIARETDB1 & 90.0\% \\ \hline
Chudzik et al. & FCNN \& Auxiliary Codebook & E-Optha MA \& E-Optha EX & & 0.8666 & 0.9998 & 0.982 \\ \hline
\end{tabular}
\end{adjustbox}
\end{table}

\subsubsection{Retinal Blood Vessel Segmentation}
Retinal blood vessels are important for different eye disease diagnosis. In this section we will discuss different results on vessel segmentation from retinal fundus images.
In one of the first studies in this domain, Maji et al. \cite{maji2015deep} used a hybrid of random forest and deep neural network (DNN) for blood vessel segmentation. The DNN performed unsupervised learning of vessel dictionaries using sparse trained denoising auto-encoders (DAE). It was followed by supervised learning of random forest on the DNN response. However this method could not outperform the conventional approaches. Around the same time Liskowski et al. \cite{liskowski2016segmenting} proposed a deep learning based blood vessel segmentation framework of retinal fundus images datasets. Images were standardized by subtracting the mean from every patch and dividing it by the standard deviation to avoid contrast and brightness fluctuations in the image pixels. It outperformed many existing approaches. In another pioneering work Melinscak et al. \cite{melinvsvcak2015retinal} used a deep neural network,  inspired by a similar problem of segmenting neuronal membranes \cite{cirecsan2012multi} using DNN as pixel classifier. To improve the performance proposed in \cite{maji2015deep} Maji et al. \cite{maji2016ensemble} used an ensemble of 12 CNNs to segment retinal blood vessels. The networks were trained individually on the dataset of 60,000 randomly chosen 3x31x31 patches. During inference, the responses were averaged to form the final segmentation. RMSProp was used as the optimizer and a minibatch size of 200 was used. Fu et al. \cite{fu2016retinal} found several disadvantages in \cite{melinvsvcak2015retinal} as it used pixel based approach and hence Fu et al. proposed a fully CNN architecture based on image-to-image training system. Multi-scale and multi-level CNN was used and combined with conditional random field (CRF) to model the long-range interactions between pixels. Leopold et al. \cite{leopold2017segmentation} investigated use of CNN to segment blood vessels using ADAM parameter optimization. The green channel of each image was used for classifying vessels and non-vessels image pixels. The model gave the probability maps of every pixel to classify between vessel and non-vessel. Gabor filters were used to smooth and finalize the decision. Zhang et al. \cite{zhang2018deep} applied U-Net which was also used in other works for OD segmentation \cite{feng2017deep} \cite{sevastopolsky2017optic}. The authors proposed a modified U-Net based architecture to segment blood vessels from fundus images. By adding some additional labels on boundary areas the problem was converted into a multi-class task.  Stochastic gradient descent (SGD) was used to optimize model parameters. Oliveira et al. \cite{oliveira2018retinal} implemented deep learning architecture for blood vessel segmentation. Previous deep learning architectures only processed raw data but here, initially, stationary wavelet transform was applied to each training image to keep multi-resolution information. A fully convolutional network was used to generate feature maps. Stochastic Gradient Descent with Nesterov momentum was implemented during training to decrease the cross-entropy loss function. The final probability maps for all of the image patches were merged and averaged to get a final value and thresholding was done to get the ultimate unique segmentation. Liu et al. \cite{liu2018retinal} used densely Connected CNN to segment blood vessels in fundus images. A 17 layer architecture was used and the layer number X got input from the output of the previous X-1 layers and thus used the back layers of the network as features of the front layer. Similar to \cite{fu2016retinal} Hu et al. \cite{mo2018exudate} proposed an image-to-image deep learning vessel detection model using the CNN combined with conditional random field (CRF). Main contribution of this work was to combine features from each of the convolutional layers and to incorporate class-balanced cross-entropy loss to improve detection accuracy. VGG-16 model was used. 
Lepetit-Aimon et al \cite{lepetit2018large} introduced the LRFFCN which did better than the U-Net \cite{ronneberger2015u}in retinal artery and vein classification and manifested high sensitivity in comparison to other state of the art algorithms to segment vessels. Chudzik et al. \cite{chudzik2018discern} gave a two stage architecture combining visual codebook framework with CNN.
\begin{table}[hbt!]
\caption {Summary of Retinal Blood Vessel Segmentation Results} \label{tab:title}
\begin{adjustbox}{width=1\textwidth}
\begin{tabular}{m{2.5cm}m{2.5cm}m{2.5cm}m{2.5cm}m{2.5cm}m{2.5cm}m{2.5cm}m{2.5cm}m{2.5cm}}
\hline
\textbf{Reference} & \textbf{Architecture} & \textbf{Dataset} & \textbf{Acc} & \textbf{SN} & \textbf{SP} & \textbf{AUC} \\ \hline
Maji et al. & RF and DNN & DRIVE & 93.27\% & & & .9195 \\ \hline
Liskowski et al. & CNN & DRIVE, STARE, CHASE\_DB & 97\% & & & .99 \\ \hline
Maji et al. & ConvNet ensemble & DRIVE & 94.7\% & & & .9283 \\ \hline
Fu et al. & Multi-scale and Multi-level CNN & DRIVE, STARE, CHASE\_DB1 & 95.23\%, 95.85\%, 94.89\% & & & \\ \hline
Leopold et al. & FCN using RETSEG13 & DRIVE & 94.78\% & 68.23\% & 98.01\% & .9707 \\ \hline
Zhang et al. & CNN(U-Net) & DRIVE, STARE, CHASE\_DB1 & 95.04\%, 97.12\%, 97.7\% & 87.23\%, 76.73\%, 76.7\% & 96.18\%, 99.01\%, 99.09\% & .9799, .9882, .99\\ \hline
Oliveira et al. & CNN & DRIVE, STARE, CHASE\_DB1 & 95.76\%, 96.94\%, 96.53\% & 80.39\%, 83.15\%, 77.79\% & 98.04\%, 98.58\%, 98.64\% & .9821, .9905, .9855 \\ \hline
Liu et al. & Densely Connected CNN & DRIVE & 95\%  \\ \hline 
Lepetit-Aimon et al. & FCNN with large receptive field & MESSIDOR, STARE, DRIVE & 95.9\%  \\ \hline 
Chudzik et al. & CNN & DRIVE, STARE & & 0.7881, 0.8269 & 0.9741, 0.9804 & 0.9646, 0.9837 \\ \hline
\end{tabular}
\end{adjustbox}
\end{table}

\subsubsection{AMD Classification}\label{sec:amd}
Recent results on AMD disease classification from fundus images are discussed below.

One of the very first publications in this domain was done by Burlina et al. \cite{burlina2016detection} where they used OverFeat features from DCNN (pretrained in ImageNet database) and used Support Vector Machine to classify between early and intermediate stages of AMD.
Later Burlina et al. \cite{burlina2017automated} used a completely data-driven approach using deep CNN (DCNN-A) to perform a binary classification between early-stage AMD and advanced stage AMD using the same AREDS database implemented on AlexNet model \cite{guo2016deep}. This method was compared with the previous methods combining both deep features and transfer learning.Based upon the work of \cite{burlina2016detection} Horta et al. \cite{horta2017hybrid} reported a hybrid method employing deep image features and random forest to combine different patient non-visual data e.g. lifestyle, cataract, demographics with the image for AMD classification. To extract deep image features, the CNN (pre-trained with 1.2 million image data) was used. The deep features combined with the non-medical, non-visual information of the patients were used to train a Random Forest Classifier to perform binary classification for higher severity AMD and lower severity of AMD. The combined features were found to achieve higher accuracy than individual feature set.
Govindaiah et al.\cite{govindaiah2018deep} reported an extended study of \cite{burlina2017comparing} with a modified deeper VGG16 architecture. The macula was chosen as a Region of Interest and images were resized to a common reference level. For comparison with the VGG16, a 50 layer Keras implementation of residual neural network was used. Matsuba et al. \cite{matsuba2018accuracy} published a new approach for detecting AMD disease from ultra wide-range Optos ophthalmoscope color fundus images. Three convolutional layers, with ReLU unit and max-pooling layers were used to perform this experiment on pre-processed fundus images. The accuracy of DCNN using images was compared with human grading by six ophthalmologists. Tan et al. \cite{tan2018age} used a 14 layer CNN to detect AMD. Three fully-connected layers, 4 max-pooling layers, and 7 convolutional layers were implemented in this work. Adam optimization \cite{kingma2014adam} was used for tuning the CNN model's parameters.Grassmann et al. \cite{abc} proposed a deep learning based classification architecture to predict the severity of AMD. In this study, an ensemble of several convolutional neural networks was used to classify among 13 different classes of AMD \cite{ying2009description}. Mainly four different steps can be found in this methodology. Six different neural networks (AlexNet, GoogLeNet, VGG, ResNet, Inception-v3, 1-ResNet-v2 ) were used independently to train the model. With the result obtained from each of the individual neural networks, a random forest ensemble model was developed.
Govindaiah et al. \cite{govindaiah2018new} used an ensemble network consisting of state of the art network architectures thereby reaching a satisfactory performance level in AMD classification.

\begin{table}[hbt!]
\caption {Summary of AMD Detection Studies} \label{tab:title}
 \begin{adjustbox}{width=1\textwidth}
\begin{tabular}{m{2.5cm}m{3cm}m{2.5cm}m{2.5cm}m{2.5cm}m{2.5cm}m{2.5cm}m{2.5cm}m{2.5cm}}
\hline
\textbf{Reference} & \textbf{ Architecture} & \textbf{Dataset} & \textbf{Acc} & \textbf{SN} & \textbf{SP} & \textbf{AUC}\\
\hline
Burlina et al. & Deep Features with SVM & AREDS & 95\% & 96.4\%& 95.6\% &\\ \hline
Burlina et al. & DCNN using AlexNet & AREDS & 91.6\%& & & .96\\ \hline
Horta et al. & DCNN & AREDS & 79.04\% & 66.34\% & 88.95\% & .8476 \\ \hline
Govindaiah et al. & VGG16 & AREDS dataset &92.5\% \\ \hline
Matsuba et al. & DCNN & Tsukazaki  Hospital &  & & & 99.76\% \\ \hline
Tan et al. & CNN & Kasturba Medical College & 95.45\% & 96.43\% & 93.45\% \\ \hline
Grassmann et al. & Ensemble (AlexNet, GoogleNet, VGG, ResNet, Inception-v3, 1-ResNet-v2) & AREDS and KORA & & 94.3\% & 84.2\%   \\ \hline
Govindaiah et al. & Ensemble Network (Inception-ResNet-V2 \& Xception) & AREDS & 86.13\%\\ \hline

\end{tabular}
\end{adjustbox}
\end{table}

\subsubsection{Glaucoma Classification}
One of the early publications in glaucoma classification using deep learning was by Chen et al.\cite{chen2015glaucoma}. They implemented a CNN with dropout and data augmentation. A six layers deep CNN with 4 convolutional layers of progressively decreasing filter size (11, 5, 3, 3) followed by 2 dense layers was used.
Improving their previous work, \cite{chen2015glaucoma} Chen et al. \cite{chen2015automatic} presented a model using Contextualized CNN (C-CNN) architecture. It combined the output of convolutional layers of multiple CNN to a final dense layer to obtain the softmax probabilities. The 5 C-CNN model which was a concatenation of outputs of last convolutional layers of 5 CNNs each of depth 6 (5 convolutional layers + 1 MLP). Asoaka et al. \cite{asaoka2016} used a 3 layer deep Feed-forward Neural Network (FNN) on a private dataset of 171 Glaucoma images. 
Chakravarty \cite{chakravarty2018} was first to propose a method for joint segmentation of OD, OC and glaucoma prediction. In this method CNN feature sharing for different tasks ensured better learning and over-fitting prevention. The parts of the model that were shared with U-net contained 8 times fewer number of CNN filters than the conventional U-net. It used an encoder network to downsample the feature and then a decoder network to restore the image size. Two different convolutional layers were applied on the decoder network's output for OC and OD segmentation. The OC and OD segmentation masks were merged into separate channels and CNN was applied to it. The outputs of the CNN and encoder output were combined and fed to a single neuron to predict glaucoma. With a lower number of parameters this method achieved comparable performance with existing architecture e.g. \cite{fu2018disc}. Zhixi et al. \cite{zhixi2018} used the Inception-v3 architecture to detect glaucomatous optic neuropathy. Here researchers graded the images by trained ophthalmologists before applying the algorithm. Local space average color subtraction was applied in pre-processing to accommodate for varying illumination. 
Chai et al.\cite{chai2018} presented a framework on a dataset of fundus images obtained from various hospitals by incorporating both domain knowledge and features learned from a deep learning model. This method was also used in other applications \cite{orlando2018ensemble}  The disk image provided local CNN features, the whole image provided global CNN features whereas domain knowledge features were obtained from diagnostic reports. It used a total of 25 features including 3 numerical features: intraocular pressure, age, and visual acuity as well as 22 binary features such as swollen eye, headache, blurred vision and failing visual acuity. The disk and whole images were fed to two separate CNN while domain knowledge features were fed to a third branch consisting of a fully connected neural network. These three branches were concatenated by a merge layer followed by two dense layers and a logistic regression classifier. 
Perdomo et al. \cite{perdomo2018glaucoma} used curriculum learning \cite{bengio2009curriculum} in DCNN's to achieve better results using a reduced set of training examples. Pal et al. \cite{pal2018g} introduced the G-EyeNet architecture which proved to be more robust given its results on low quality images.
\begin{table}
\caption {Summary of Glaucoma Detection Studies} \label{tab:title}
\begin{adjustbox}{width=1\textwidth}
\begin{tabular}{m{2.5cm}m{2.5cm}m{2.5cm}m{2.5cm}m{2.5cm}m{2.5cm}m{2.5cm}m{2.5cm}m{2.5cm}m{2.5cm}}
\hline
\textbf{Reference} & \textbf{Architecture} & \textbf{Dataset} & \textbf{Acc} & \textbf{SN} & \textbf{SP} & \textbf{AUC} & \textbf{F-Score}\\ \hline
Chen et al. &  6 layer CNN & ORIGA, SCES && & & .831, .887 \\ \hline
Asoaka et al. & 3 layer FNN & Private: 171& & & &.926 \\ \hline
Chakravarty et al. &  Multi-task CNN & REFUGE &&&& .9456 \\ \hline
Zhixi et al. & Inception-v3 & Private:48000+ & & 95.6\%& 92.0\%& .986\\ \hline
Chai et al. & MB-NN & Private: 2554 &91.51\% & & & \\ \hline
Chen et al. & C-CNN & ORIGA, SCES && & & .838, .898\\ \hline
Perdomo et al. & DCNN & RIM-ONE-v1, RIM-ONE-v3, DRISHTI-GS1 & 89.4\%(RIM-ONE-v1) & & & 0.82 (DRISHTI-GS) \\ \hline
Pal et al. & CAE with CNN classifier & DRIONS-DB & & & & 0.923 \\ \hline

\end{tabular}
\end{adjustbox}

\end{table}

\subsubsection{Diabetic Retinopathy Classification}
In this section different applications of deep learning algorithms for diabetic retinopathy detection are described briefly.

Abràmoff et al \cite{abramoff2016improved} described DR detection using a device called IDx-DR X2.1. Here retinal images were used in a CNN based on AlexNet to classify different types of DR. The main classes of diseases were referable DR (rDR), vision-threatning DR (vtDR) and proliferative DR (pDR). The CNN-based architectures were designed to characterize and detect optic disc, fovea and lesion characteristics.
Using a retinal fundus image dataset consisting of 70000 images, Colas et al. \cite{colas2016deep} proposed a DR grading method. There were 4 different classes of DR images in the dataset- no DR, mild DR, moderate DR and acute DR. Gulshan et al. \cite{gulshan2016development} used deep learning algorithms to identify the presence of diabetic retinopathy. Five different types of DR and the presence of macular edema were graded by expert clinicians. Inception v-3 model was used, stochastic gradient descent method for optimization was used and batch normalization was done with a pre-trained model with ImageNet data. Gargeya et al.\cite{gargeya2017automated} reported a deep learning architecture to classify between normal and DR fundus images and also reported heatmap visualization of the result. Using the principle of deep residual learning, the CNN model was built to learn deep discriminative features for detecting DR. From average pooling layer of CNN, 1024 features were obtained. Metadata features related to 3 metadata variables i.e. pixel height, pixel width and field of view of the image were appended to form a final feature vector with 1027 features. A second level tree-based gradient boosting classifier was designed. Quellec et al. \cite{quellec2017deep} discussed a method to detect referable DR as well as lesions with ConvNet architecure using o\_O Solution \cite{kagg}.Unlike the previous studies this method attempted to classify between normal and DR on both image level and pixel level. This proposed model was mainly based on visualization methods of CNN. Heatmap generation modifications were proposed for this purpose to jointly improve the quality of DR and lesion detection. Takahashi et al. \cite{takahashi2017applying} graded different stages of the presence of DR using GoogleNet architecture. Unlike in other previously published literature the authors graded the images manually on their own to test the accuracy of the methodology. The model was designed using two different ways, first with manual staging of three color photographs (AI1) and second with manual staging of only one color photograph (AI2). From the GoogleNet model- 5 top layers were deleted, the crop size was expanded and the batch size was reduced. 20 fold cross-validation was used and for comparison AI1 was also trained with ResNet model. García et al. \cite{garcia2017detection} applied different architectures of CNN for DR detection. As a pre-processing step, images were subtracted from color mean and rescaled to 256x256. Data augmentation by flipping the images was done to increase the robustness. In this work several neural network architectures using various learning rates and different number of layers were used to compare different architectures to calculate the highest accuracy among all. Lin et al. \cite{lin2018transforming} used entropy images instead of original fundus images and showed that the feature maps are generated faster and competently.
\begin{table}[hbt!]
\caption {Summary of Diabetic Retinopathy Detection Studies} \label{tab:title}
\begin{adjustbox}{width=1\textwidth}
\begin{tabular}{m{2.5cm}m{3cm}m{2.5cm}m{2.5cm}m{2.5cm}m{2.5cm}m{2.5cm}m{2.5cm}m{2.5cm}}
\hline
\textbf{Reference} & \textbf{Architecture} & \textbf{Dataset} & \textbf{Acc} & \textbf{SN} & \textbf{SP} & \textbf{AUC} \\
\hline
Abràmoff et al. &  CNN AlexNet & Messidor-2 & & 96.8\% & 87\% \\ \hline
Colas et al. & CNN & EyePACS & & 96.2\% & 66.6\% & .946\\ \hline
Gulshan et al. & CNN Incetionv-3 & EyePACS-1, Messidor-2 & & 90.3\%, 87\% & 98.1\%, 98.5\% & .991, .99 \\ \hline
Gargeya et al. & CNN Deep Residual Learning & EyePACS, MESSIDOR e-Optha2 & & 94\% & 98\% & .97 \\ \hline
Quellec et al. & ConvNet & Kaggle, e-optha, DiaretDB1 & & & & .954, .949, .955 \\ \hline
Takahashi et al. & CNN GoogleNet, ResNet & 9939 images & 80\% \\ \hline
Garcia et al. & CNN AlexNet & EyePACS & 83.68\% & & \\ \hline
Lin et al. & CNN & Kaggle & 86.10\% & 73.24\% & 93.81\% & 0.92 \\ \hline

\end{tabular}
\end{adjustbox}
\end{table}
\section{Conclusion and Future Research} \label{sec:disc}
This review addressed different applications of deep learning methodologies in ophthalmic diagnosis.
\begin{table}[h!]
\caption {Deep Learning vs Traditional Methods} \label{tab:title}
\begin{adjustbox}{width=1\textwidth}
\begin{tabular}{m{2.5cm}m{2.5cm}m{2.5cm}m{2.5cm}m{2cm}m{2cm}m{2cm}m{2cm}m{2cm}}
\hline
\textbf{Application} & \textbf{Reference} & \textbf{Method} & \textbf{Dataset} & \textbf{Acc} & \textbf{SN} & \textbf{SP} & \textbf{AUC} & \textbf{F1 Score} \\
\hline
OD Segmentation & Maninis et al. \cite{maninis2016deep} & CNN DRIU & DRIVE, STARE & & & & & .822, .831 \\
& Soares et al. \cite{soares2006retinal} & Wavelets & DRIVE, STARE & & & & & .762, .774 \\ \hline

Lesion Detection & Haloi et al. \cite{haloi2015improved} & 5-layer CNN & Messidor & 96\% & 97\% & 96\% & .988 &  \\
 & Antal et al. \cite{antal2014ensemble} & Ensemble Model & Messidor & 90\% & 90\% & 91\% & .989 \\  \hline

Retinal Vessel Segmentation & Leopold et al. \cite{leopold2017segmentation}& FCN using RETSEG13 & DRIVE & 94.78\% & & & & \\ 
& Staal et al. \cite{staal2004ridge} &  kNN Classifier & DRIVE & 94.22\% & & & & \\ \hline

AMD Classification & Burlina et al. \cite{burlina2016detection}& AREDS & Deep Features with SVM &  95\% & 96.4\% & 95.6\% & & \\
 & Kankanahalli et al. \cite{10.1167/iovs.12-10928} & AREDS & SURF features with random forest &  91.8\% & 91.3\% & 92.3\% \\ \hline

Glaucoma Classification & Perdomo et al. \cite{perdomo2018glaucoma}& DCNN & RIM-ONE-v1, RIM-ONE-v3, DRISHTI-GS1 & 89.4\%& & &  & \\
 & Gajbhiye et al. \cite{gajbhiye2015automatic} & KNN & & 89\% & & & &  \\ \hline

DR Classification & Gargeya et al. \cite{gargeya2017automated} & CNN Deep Residual Learning & EyePACS, MESSIDOR e-Optha2 & & 94\% & 98\% & .97 & \\
  & Roychowdhury et al. \cite{roychowdhury2013dream} & kNN (DREAM) & MESSIDOR & & 98.88\% & 48.72\% & & \\
\hline
\end{tabular}
\end{adjustbox}
\end{table}
Table 8 gives a brief overview of state-of-the-art deep learning approach and traditional imethods for computer-aided diagnosis.It can be noticed that in most of the cases deep learning methods outperformed traditional methodologies.\\
The previous reviews published in this domain were very clinical or focused on traditional machine learning algorithms or emphasized a particular disease or focused on hardware implementation of artificial intelligence in ophthalmic diagnosis \cite{zhang2014survey}\cite{teikari2018embedded}\cite{rahimy2018deep}\cite{hogarty2018current}\cite{salamat2018diabetic}. None of them dealt with detailed reviews of different state-of-the art deep learning algorithms used in ophthalmic diagnosis with retinal fundus images. Hence, to the best of our knowledge this is the first review article of deep learning algorithms and performance outcomes for different architectures for ophthalmic diagnosis using retinal fundus images.Deep learning applications in retinal images are quite useful and effective. They reduce the need of manual feature extraction as the methodologies are mainly data-driven. Convolutional neural networks  are the most widely used architecture for classification, detection or segmentation of different parts of fundu images. Ensemble, FCN, ResNet and AE based architectures were commonly used in these studies.
However, there are still some limitations which need to be addressed. Some of these and also some possible solutions are discussed below: 
\begin{itemize}
    \item Unlike computer vision problems, large datasets are not available. Also there is a scarcity of manual annotation of data. Deep learning equated large amounts of data since the model mainly learns from the inherent pattern of the data. Hence this is a major problem in this field.Generative models proposed by Goodfellow et al. can be an important and useful solution to mitigate this problem. This is a very state-of-the art area of this research and very few efforts have been made so far \cite{diaz2019retinal} \cite{costa2017towards} to explore the possibilities of generative modelling to synthesize new fundus images with annotations and with proper clinical relevance. Generative Adversarial Network, Variational Auto-encoders are some very popular architecture for image generation. Successful application of these can be used to generate large amounts of clinically relevant synthetic data. It will not only help to increase amount of available data but also it will help to avoid the privacy issues.
    \item A major problem is the unavailability of standardized KPIs (Key Performance Indicators) for measuring the performance of a particular model. Different researchers use different indices to measure their work. Due to this variablity one cannot easily compare different deep learning architectures for a given disease state. For example, in lesion detection, Lam et al. \cite{lam2018retinal} achieved an accuracy of 98\% which is higher than most of the other state-of-the-art methods, whereas in terms of AUC Haloi\cite{haloi2015improved} achieved 0.982 which is higher than other reported AUC. Leopold et al. \cite{leopold2017pixelbnn} took this into consideration and also suggested more generalized metrics such as G-mean and MCC to measure a model's effectiveness.
    \item Due to the difference in camera settings there is a possibility of domain shift problem. In most of the literature, training and test data come from same image distribution. But in real life this is not always the case. Hence this domain shift can cause a major damage in real life application if not taken care of beforehand. Transfer learning has been used for different applications in this area \cite{edupuganti2018automatic} \cite{burlina2017automated}\cite{kermany2018identifying}\cite{chan2017transfer}. Domain Adaptation is a sub-domain of Transfer Learning where data for both training and testing are extracted from different distributions. In real world, it is not always possible to get test data and training data from the same distribution. Hence the model should be robust enough to deal with data from a different distribution for test purpose. Often it is found that accuracy decreases due to this domain shift problem. More emphasis should be given to deep domain adaptation approaches in order to create robust models which can be implemented for real world ophthalmic diagnosis. Wang et al. \cite{wang2018deep} have discussed different deep domain adaptation algorithms which can be used to address this problem. A recent paper explored adversarial domain adaptation technique to segment blood vessels of STARE dataset with a model trained on DRIVE dataset and it outperformed other works in terms of F score \cite{javanmardi2018domain}. In the context of ophthalmic diagnosis it can be an important and necessary direction for future research.

\end{itemize}

 \section{Acknowledgement}
This research was supported by a Discovery Grant from NSERC, Canada to Vasudevan Lakshminarayanan.
 \section{Conflict of Interest}
 The authors declare no conflict of interest.



\bibliographystyle{unsrt}
\bibliography{main}
\end{document}